\documentclass[10pt,twocolumn,letterpaper]{article}

\usepackage{cvpr}
\usepackage{times}
\usepackage{epsfig}
\usepackage{graphicx}
\usepackage{amsmath}
\usepackage{amssymb}
\usepackage{breqn}
\usepackage{tabularx}
\usepackage{multirow}
\usepackage{color}
\usepackage{lineno}
\usepackage{subfigure}
\usepackage{epstopdf}

\usepackage[pagebackref=true,breaklinks=true,letterpaper=true,colorlinks,bookmarks=false]{hyperref}

\cvprfinalcopy % *** Uncomment this line for the final submission

\ifcvprfinal\fi
\begin{document}

%%%%%%%%% TITLE
\title{Multi-scale Location-aware Kernel Representation for Object Detection}

\author{Hao Wang$^{1}$, Qilong Wang$^{2}$,  Mingqi Gao$^{1}$,  Peihua Li$^2$, Wangmeng Zuo$^{1,}$\thanks{Corresponding author.  \newline\indent{ This work is supported by NSFC grants (No.s 61671182, 61471146, and 61471082).}}\\
$^1$School of Computer Science and Technology, Harbin Institute of Technology, Harbin, China\\
$^2$School of Information and Communication Engineering, Dalian University of Technology, Dalian, China\\
{\tt\small ddsywh@yeah.net, qlwang@mail.dlut.edu.cn, hit.gmq@gmail.com }\\
{\tt\small peihuali@dlut.edu.cn, wmzuo@hit.edu.cn}
}

\maketitle
\thispagestyle{empty}

%%%%%%%%% ABSTRACT
\begin{abstract}
  Although Faster R-CNN and its variants have shown promising performance in object detection, they only exploit simple first-order representation of object proposals for final classification and regression. Recent classification methods demonstrate that the integration of high-order statistics into deep convolutional neural networks can achieve impressive improvement, but their goal is to model whole images by discarding location information so that they cannot be directly adopted to object detection. In this paper, we make an attempt to exploit high-order statistics in object detection, aiming at generating more discriminative representations for proposals to enhance the performance of detectors. To this end, we propose a novel Multi-scale Location-aware Kernel Representation (MLKP) to capture high-order statistics of deep features in proposals. Our MLKP can be efficiently computed on a modified multi-scale feature map using a low-dimensional polynomial kernel approximation. Moreover, different from existing orderless global representations based on high-order statistics, our proposed MLKP is location retentive and sensitive so that it can be flexibly adopted to object detection. Through integrating into Faster R-CNN schema, the proposed MLKP achieves very competitive performance with state-of-the-art methods, and improves Faster R-CNN by 4.9\% (mAP), 4.7\% (mAP) and 5.0\% (AP at IOU=[0.5:0.05:0.95]) on PASCAL VOC 2007, VOC 2012 and MS COCO benchmarks, respectively. Code is available at: \url{ https://github.com/Hwang64/MLKP}.

  %Despite their popularity in object detection, Faster R-CNN and its existing variants all only exploit simple first-order representation of object proposals for final classification and regression.
  %While high-order statistics with deep features have shown the ability in improving classification performance, most existing methods ignore the spatial location of feature maps and cannot be directly adopted in object detection.
  %To address above limitation, this paper proposes a novel Multi-scale Location-aware Kernel Representation (MLKP) to capture high-order statistics of features in proposals.
  %To our best knowledge, our MLKP makes the first attempt to exploit high-order statistics in object detection to generate more discriminative representations than commonly used first-order ones.
  %Although polynomial kernel approximation based high-order statistics have successfully applied to classification, different from their orderless global representations, our proposed MLKP is location retentive and sensitive so that it can be flexibly adopted to object detection.
  %Our MLKP can naturally incorporate multi-scale information, and can be conveniently integrated into Faster R-CNN framework to significantly improve performance of object detection. Specifically, it improves Faster R-CNN by $4.9\%$ (mAP), $4.7\%$ (mAP) and $5.0\%$ (AP at IOU=[0.5:0.05:0.95]) on PASCAL VOC07, VOC12 and MS COCO benchmarks, respectively. Finally, the comprehensive experiments show combination of our MLKP with Faster R-CNN is very competitive with state-of-the-art methods.

\end{abstract}

%%%%%%%%% BODY TEXT
\section{Introduction}

\begin{figure}[!t]
  \scriptsize{
  \begin{center}
  \subfigure[Faster R-CNN~\cite{ren2015faster}]
  {\includegraphics[width=4.05cm,height=3.1cm]{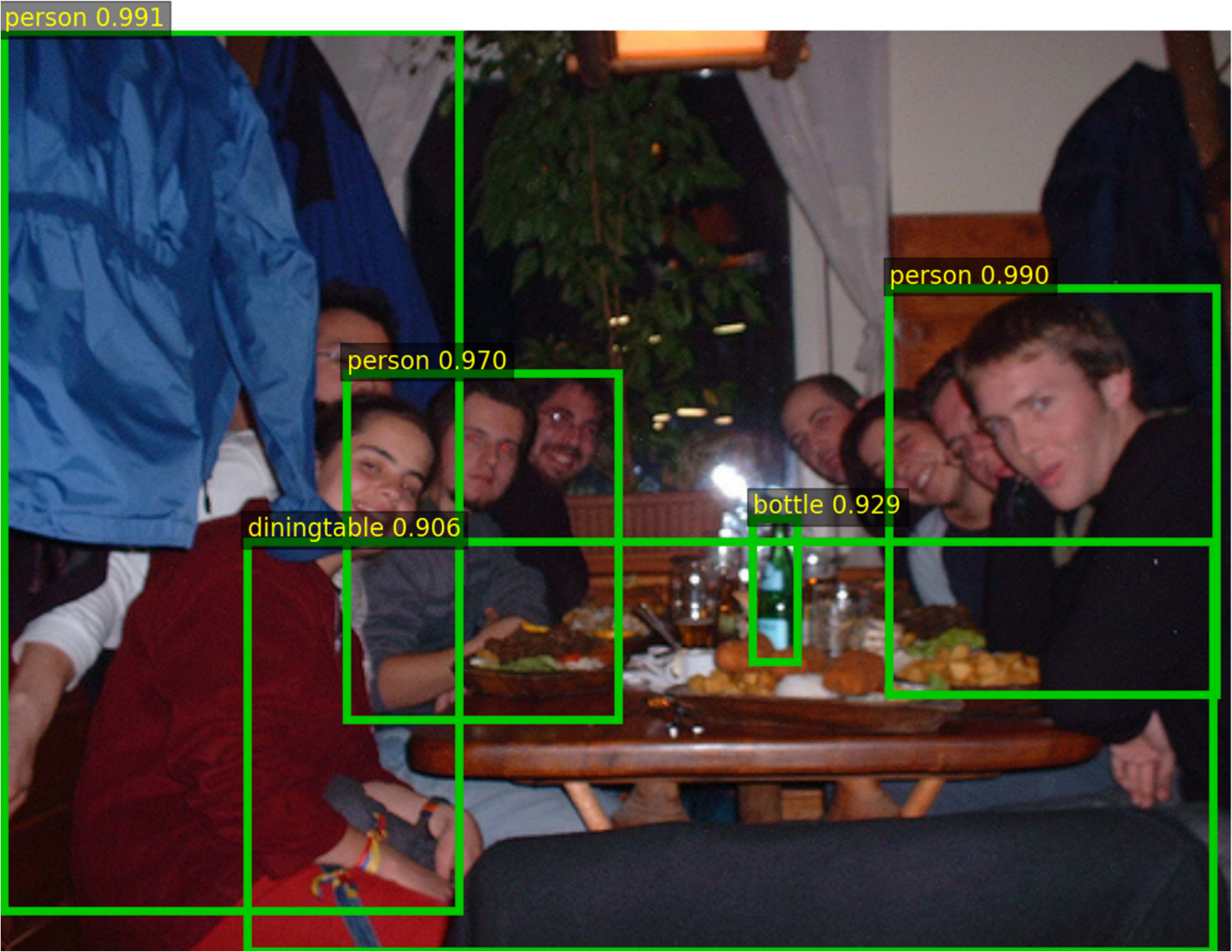}}
  \subfigure[HyperNet~\cite{kong2016hypernet}]
  {\includegraphics[width=4.10cm,height=3.1cm]{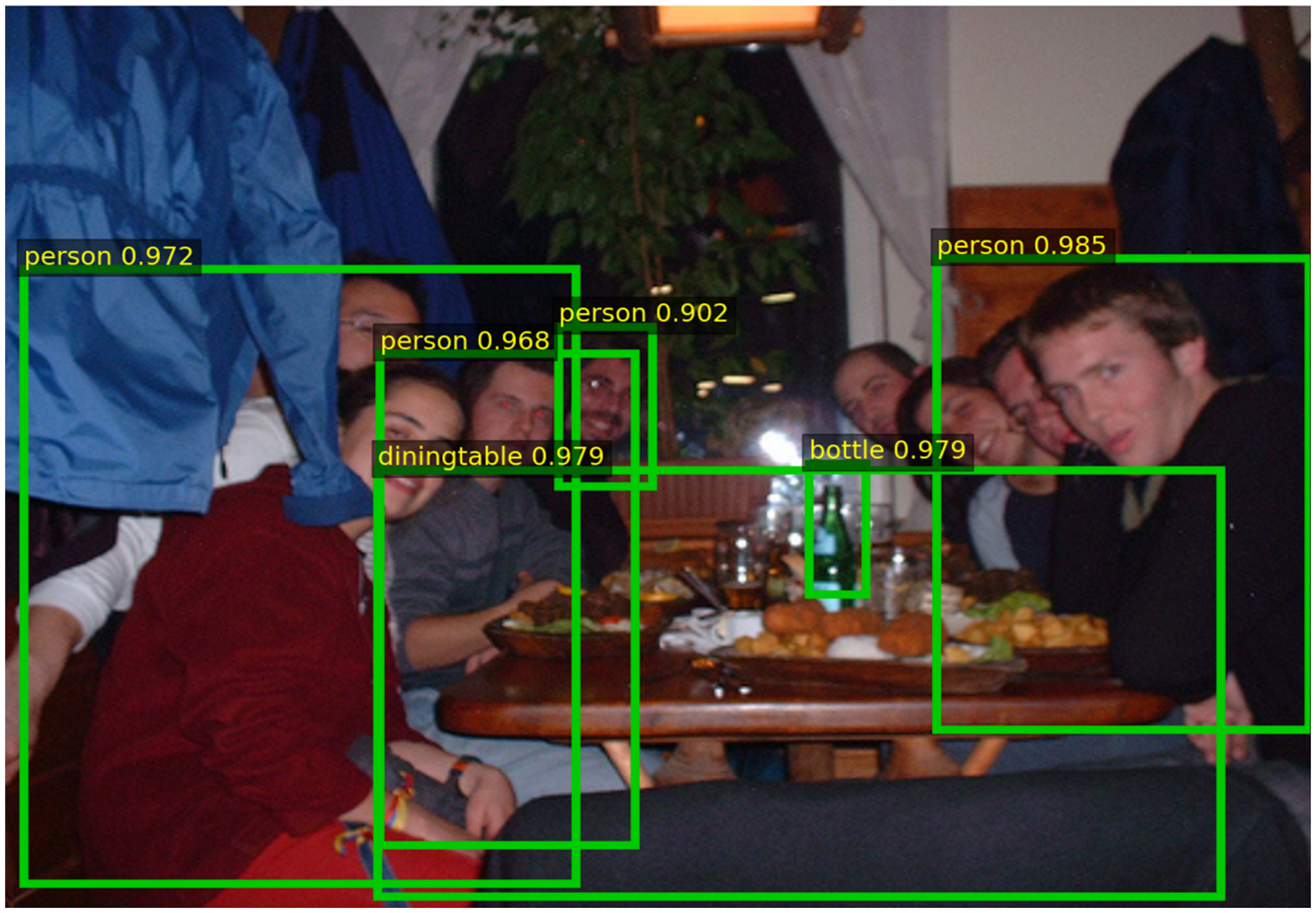}}
  \subfigure[RON~\cite{Kong_Tao_2017_CVPR}]
  {\includegraphics[width=4.05cm,height=3.1cm]{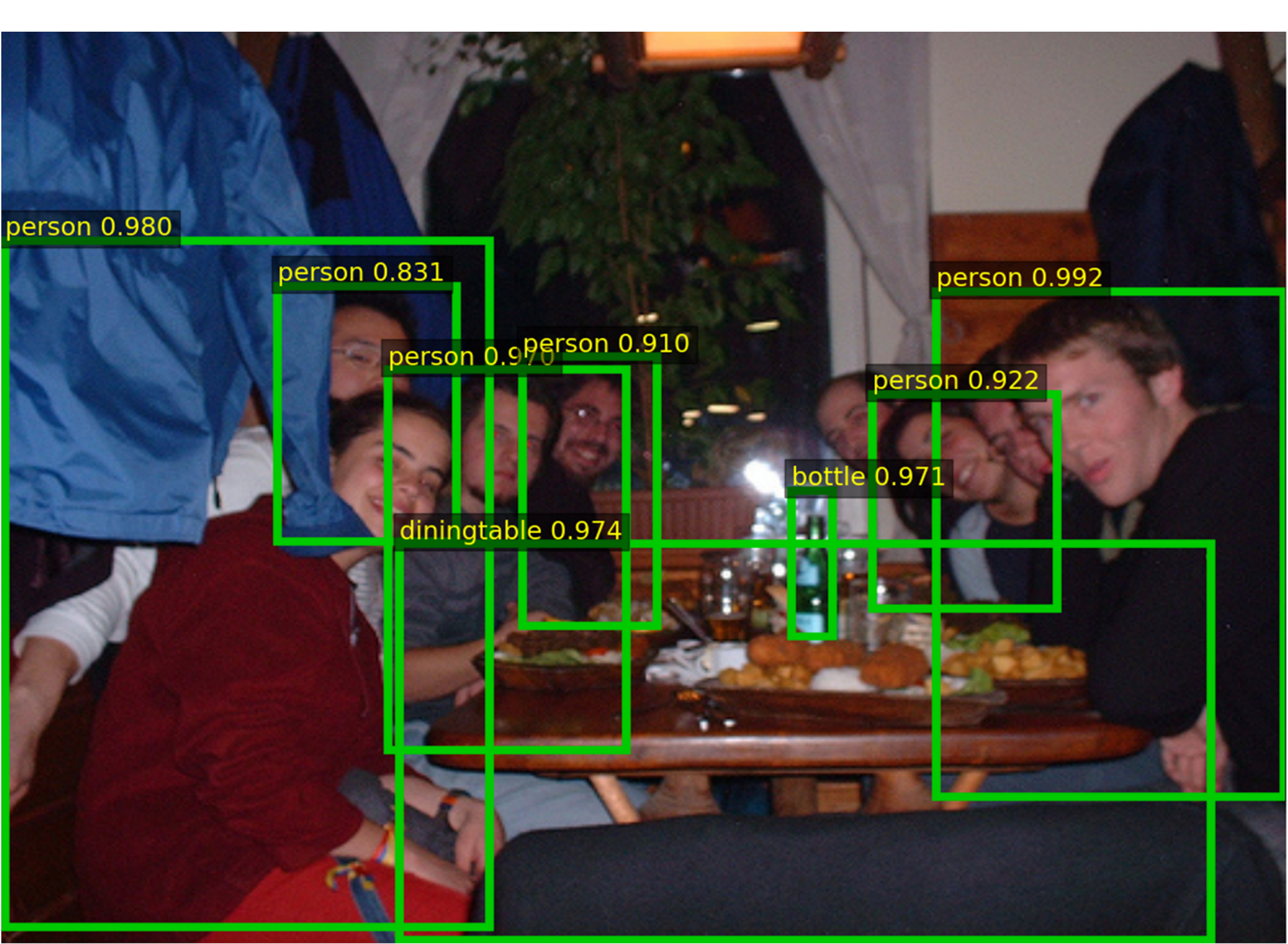}}
  \subfigure[Our MLKP]
  {\includegraphics[width=4.10cm,height=3.1cm]{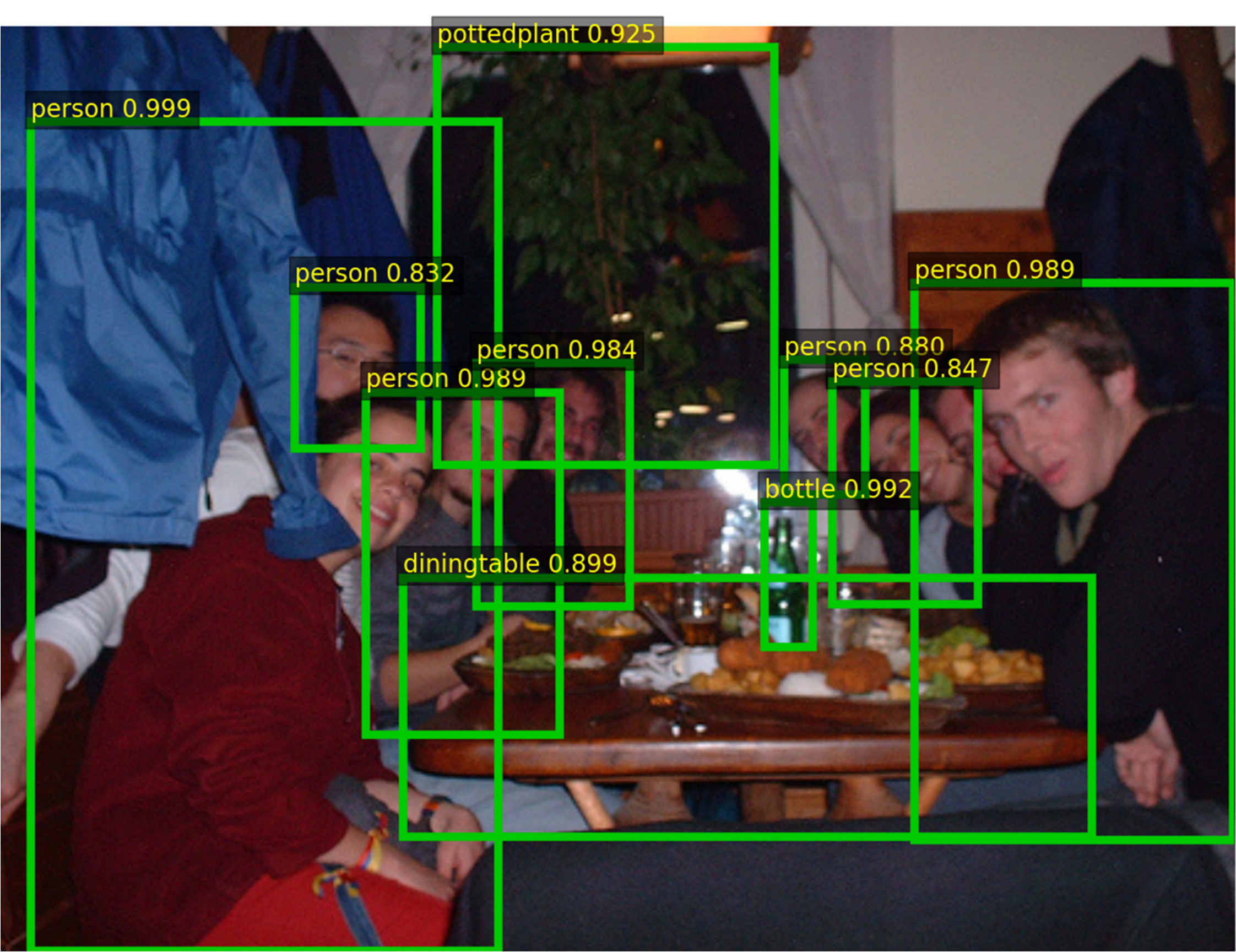}}
  \vspace{0.1cm}
  \caption{Comparison of our MLKP with Faster R-CNN and its two variants (\ie, HyperNet~\cite{kong2016hypernet} and RON~\cite{Kong_Tao_2017_CVPR}). Clearly, (a) Faster RCNN operated on a single convolutional layer locates inaccurately and mis-locates many small and hard objects. (b) HyperNet and (c) RON improve the detecting results, but they both still mis-locate persons far away camera, who have small faces. Furthermore, these three methods mis-locate the plant, which is very similar to background so that it is difficult to detect. (d) Our MLKP performs favorably in both location and classification for all objects in this image due to its more discriminative location-aware representations. Best viewed in color.}\label{fig1}
  \end{center}}\vspace{-.7cm}
\end{figure}

\begin{figure*}[!t]
  \begin{center}
  \centering\includegraphics[width=16.5cm,height=6.2cm]{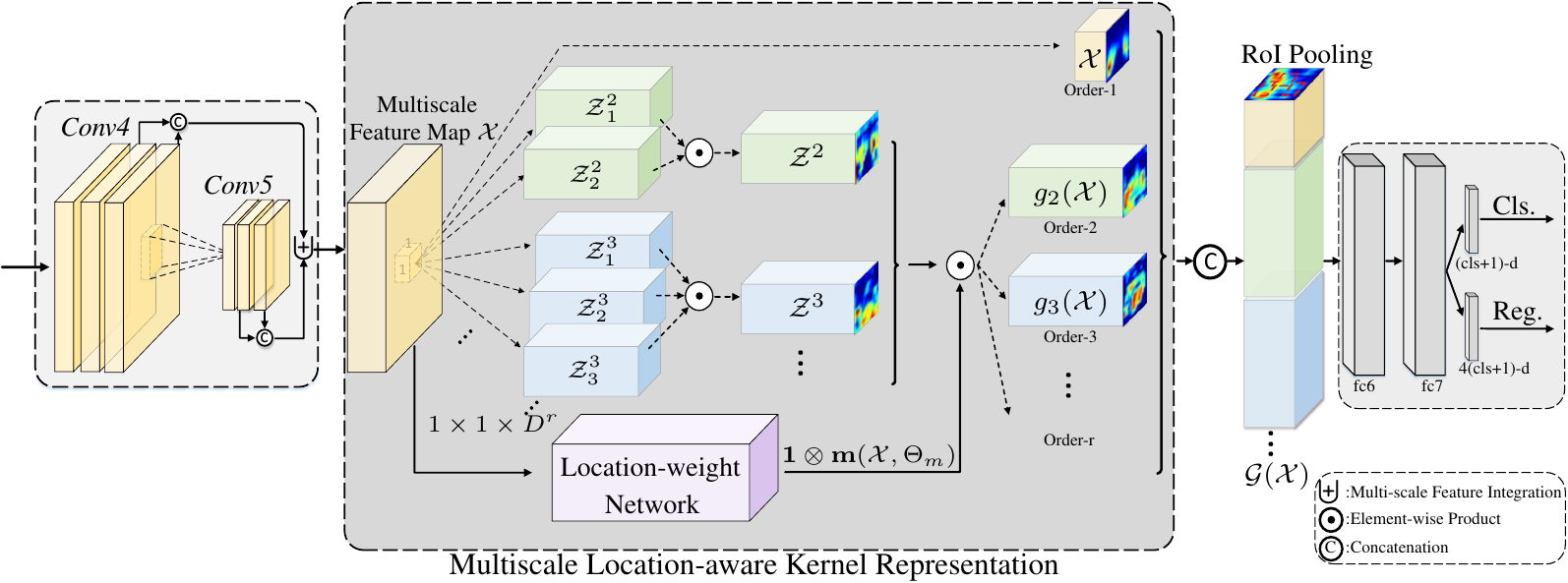}\\
  \vspace{0.1cm}
  \caption{Overview of our multi-scale location-aware kernel representation (MLKP). As polynomial kernel representations can be decomposed into $1\times1\times D^r$ convolution operations and element-wise product $\odot$. Given a multi-scale feature map $\mathcal X$ (see Fig.~\ref{fig3} for details on multi-scale feature integration $\uplus$), our MLKP first performs convolution operations and element-wise product to compute $r$-th order representation $\mathcal Z^r$, then a location-weight network $\mathbf m$ with parameter $\Theta_m$ and a remapping operation $1\otimes$ (see Fig.~\ref{fig4} for details) is learned to measure contribution of each location. The outputs of all orders polynomial kernel approximation are concatenated to generate final representation (see Eqn.~\eqref{eq6}). After that, location information still be maintained so that a commonly used RoI pooling can be adopted for final classification and regression.}\label{fig2}
  \end{center}\vspace{-0.7cm}
\end{figure*}

Object detection is one of the most fundamental and popular topics in computer vision community, and it has attracted a lot of attentions in past decades. The fast and effective object detection method plays a key role in many applications, such as autonomous driving~\cite{chen2015deepdriving}, surgical navigation~\cite{gui2013image} and video surveillance~\cite{wang2014scene}. With the rapid development of deep convolutional neural networks (CNNs)~\cite{simonyan2014very,he2016deep,szegedy2017inception}, the performance of object detection has been significantly improved. R-CNN~\cite{girshick2014rich} is among the first which exploits the outputs of deep CNNs to represent the pre-generated object proposals. R-CNN greatly improves traditional DPM~\cite{felzenszwalb2010object} and its variants~\cite{azizpour2012object}, where hand-crafted features are employed. Going beyond R-CNN, Fast R-CNN~\cite{girshick2015fast} introduces a Region of Interest (RoI) pooling layer to generate representations of all object proposals on feature map with only one CNN pass, which avoids passing separately each object proposal through deep CNNs, leading much faster training/testing process. Furthermore, Faster R-CNN~\cite{ren2015faster} designs a region proposal network (RPN) for learning to generate proposals instead of using pre-generated proposals with traditional methods~\cite{zitnick2014edge,uijlings2013selective,arbelaez2014multiscale}. By combining RPN with Fast R-CNN network (FRN), Faster R-CNN develops a unified framework by end-to-end learning. Faster R-CNN has shown promising performance in object detection, and has become a strong baseline due to its good trade-off between effectiveness and efficiency~\cite{Huang_2017_CVPR}.

Subsequently, numerous methods~\cite{Lin_2017_CVPR,Kong_Tao_2017_CVPR,bell2016inside,hariharan2015hypercolumns,Ke_2017_CVPR,kong2016hypernet,shrivastava2016beyond} have been suggested to further improve Faster R-CNN, and these methods mainly focus on one issue: original Faster R-CNN only exploits the feature map from single convolution ($conv$) layer (\ie, the last layer), leading to discard information of different resolutions, especially for small objects. As illustrated in Fig.~\ref{fig1} (a), Faster R-CNN fails to detect some small objects such as persons far away camera. There are two research directions to solve this problem, \ie, feature map concatenation~\cite{bell2016inside,hariharan2015hypercolumns,Ke_2017_CVPR,kong2016hypernet,shrivastava2016beyond} and pyramidal feature hierarchy~\cite{Lin_2017_CVPR,Kong_Tao_2017_CVPR}. The methods based on concatenation obtain a coarse-to-fine representation for each object proposal by
concatenating outputs of different convolution layers (\eg, $conv1\_3$, $conv3\_3$ and $conv5\_3$ of VGG-16~\cite{simonyan2014very} in HyperNet~\cite{kong2016hypernet}) into one single feature map. For pyramidal feature hierarchy based methods, they combine the outputs of different convolution layers in a pyramid manner (\eg, $conv5\_3$, $conv5\_3+conv4\_3$ and $conv5\_3+conv4\_3+conv3\_3$~\cite{Lin_2017_CVPR}). Moreover, each combination gives its own prediction, and all detection results are fused by using non-maximum suppression. As shown in Fig.~\ref{fig1} (b) and (c), methods based on concatenation (\eg, HyperNet~\cite{kong2016hypernet}) and pyramidal feature hierarchy (\eg, RON~\cite{Kong_Tao_2017_CVPR}) both can improve the performance by using features of different resolutions.

Although aforementioned methods can improve the detection accuracy over Faster R-CNN, they all only exploit the first-order statistics of feature maps to represent object proposals in RoI pooling. The recent researches on challenging fine-grained visual categorization~\cite{Cai_2017_ICCV,Kong_Shu_2017_CVPR,Wang_2017_CVPR,cui2017kernel} show that high-order statistics representations can capture more discriminative information than first-order ones, and obtain promising improvements. Based on this observation, we propose a Multi-scale Location-aware Kernel Representation (MLKP) to incorporate high-order statistics into RoI pooling stage for effective object detection. As illustrated in Fig.~\ref{fig1}, the method based on concatenation~\cite{kong2016hypernet} and pyramidal feature hierarchy based on~\cite{Kong_Tao_2017_CVPR} mis-locate the occluded persons far away camera. Furthermore, these methods mis-locate the plant, which is very similar to the background so that it is difficult to detect. Owing to usage of more discriminative high-order statistics, our MLKP predicts more accurate locations for all objects.

Fig.~\ref{fig2} illustrates the overview of our proposed MLKP. Through modifying the multi-scale strategy in ~\cite{hariharan2015hypercolumns}, we exploit features of multiple layers in different convolution blocks, and concatenate them into one single feature map.  Then, we compute the high-order statistics on such feature map. It is well known that the dimension of statistical information in general is $d^{k}$, where $d$ is dimension of features and $k$ denotes order number of statistical information. In our case, $d$ is usually very high (\eg, 512 or 2048 in~\cite{simonyan2014very,he2016deep}), which results in much higher dimensional representations~\cite{lin2015bilinear,Wang_2017_CVPR} and suffering from high computation and memory costs. To overcome this problem, we adopt polynomial kernel approximation based high-order methods~\cite{Cai_2017_ICCV}, which can efficiently generate low-dimensional high-order representations. To this end, the kernel representation can be reformulated with $1\times1$ convolution operation followed by element-wise product. Going beyond high-order kernel representation, we introduce a trainable location-weight structure to measure contribution of different locations, making our representation location sensitive. Finally, the different orders of representations are concatenated for classification and regression. Note that instead of global average pooling in~\cite{Cai_2017_ICCV}, we utilize max RoI pooling proposed in ~\cite{ren2015faster}, which is more suitable for object detection. To sum up, our MLKP is a kind of multi-scale, location aware, high-order representation designed for effective object detection.

We evaluate the proposed MLKP on three widely used benchmarks, \ie, PASCAL VOC 2007, PASCAL VOC 2012~\cite{everingham2010pascal} and MS COCO~\cite{lin2014microsoft}. The contributions of this paper are summarized as follows:

%\begin{enumerate}
  \noindent(1) We propose a novel Multi-scale Location-aware Kernel Representation (MLKP), which to our best knowledge, makes the first attempt to incorporate discriminative high-order statistics into representations of object proposals for effective object detection.

  \noindent(2) Our MLKP is based on polynomial kernel approximation so that it can efficiently generate low-dimensional high-order representations. Moreover, the properties of location retentive and sensitive inherent in MLKP guarantee that it can be flexibly adopted to object detection.

  \noindent(3) The experiments on three widely used benchmarks demonstrate our MLKP can significantly improve performances than original Faster R-CNN, and performs favorably in comparison to the state-of-the-art methods.
%\end{enumerate}

%--------------------------------------------------------------------------
\section{Related Work}

In contrary to region-based detection methods (\eg, Faster R-CNN and its variants), alternative research pipeline is designing region-free detection methods. Among them, YOLO~\cite{redmon2016you,Redmon_2017_CVPR} and SSD~\cite{fu2017dssd,liu2016ssd} are two representative methods. YOLO~\cite{redmon2016you} utilizes one single neural network to predict bounding boxes and class probabilities from the full images directly, which trains the network with a loss function in term of detection performance. Different from YOLO, SSD~\cite{liu2016ssd} discretizes the space of prediction of bounding boxes into a set of default boxes over several specific convolution layers. For inference, they compute the scores of each default box being to different object categories. Although region-free methods have faster training and inference speed than region-based ones, these methods discard generation of region proposals so that they often struggle with small objects and cannot filter out the negative samples belonging to the background. Furthermore, experimental results show our method can obtain higher accuracy than state-of-the-art region-free detection methods (See Sec.~\ref{section4.3} for more details). Note that it is indirect to incorporate our MLKP into region-free methods, where no object proposals can be represented by MLKP. This interesting problem is worth to be investigated in future.

Recent works have shown that the integration of high-order statistics with deep CNNs can improve classification performance~\cite{Cai_2017_ICCV,cui2017kernel,ionescu2015matrix,lin2015bilinear,Li_2017_ICCV,Wang_2017_CVPR}. Thereinto, the global second-order pooling methods~\cite{ionescu2015matrix,Li_2017_ICCV,lin2015bilinear} are plugged into deep CNNs to represent whole images, in which the sum of outer product of convolutional features is firstly computed, then element-wise power normalization~\cite{lin2015bilinear}, matrix logarithm normalization~\cite{ionescu2015matrix} and matrix power normalization~\cite{Li_2017_ICCV} are performed, respectively. Wang \etal.~\cite{Wang_2017_CVPR} embed a trainable global Gaussian distribution into deep CNNs, which exploits first-order and second-order statistics of deep convolutional features. However, all these methods generate very high dimensional orderless representations, which can not be directly adopted to object detection. The methods in~\cite{Cai_2017_ICCV,cui2017kernel} adopt polynomial and Gaussian RBF kernel functions to approximate high-order statistics, respectively. Such methods can efficiently generate low-dimensional high-order representations. However, different from methods that are designed for whole image classification, our MLKP is location retentive and sensitive to guarantee that it can be flexibly adopted to object detection.

%-------------------------------------------------------------------------
\section{Proposed Method}

In this section, we introduce the proposed Multi-scale Location-aware Kernel Representation (MLKP). Firstly, we introduce a modified multi-scale feature map to effectively utilize multi-resolution information. Then, a low-dimensional high-order representation is obtained by polynomial kernel function approximation. Furthermore, we propose a trainable location-weight structure incorporated into polynomial kernel function approximation, resulting in a location-aware kernel presentation. Finally, we show how to apply our MLKP to object detection.

\subsection{Multi-scale Feature Map}\label{sec3.1}

The original Faster R-CNN only utilizes the feature map of the last convolution layer for object detection. Many recent works~\cite{bell2016inside,hariharan2015hypercolumns,kong2016hypernet,shrivastava2016beyond} show feature maps of the former convolution layers have higher resolution and are helpful to detect small objects. These methods demonstrate that combining feature maps of different convolutional layers can improve the performance of detection. The existing multi-scale object detection networks all exploit feature maps of the last layer in each convolution block (\eg, layers of $conv5\_3$, $conv4\_3$ and $conv3\_3$ in VGG-16~\cite{simonyan2014very}). Although more feature maps usually bring more improvements, they suffer from higher computation and memory costs, especially for those layers that are closer to inputs.

\begin{figure}[!t]
\begin{center}
  \includegraphics[width=8.cm,height=6.5cm]{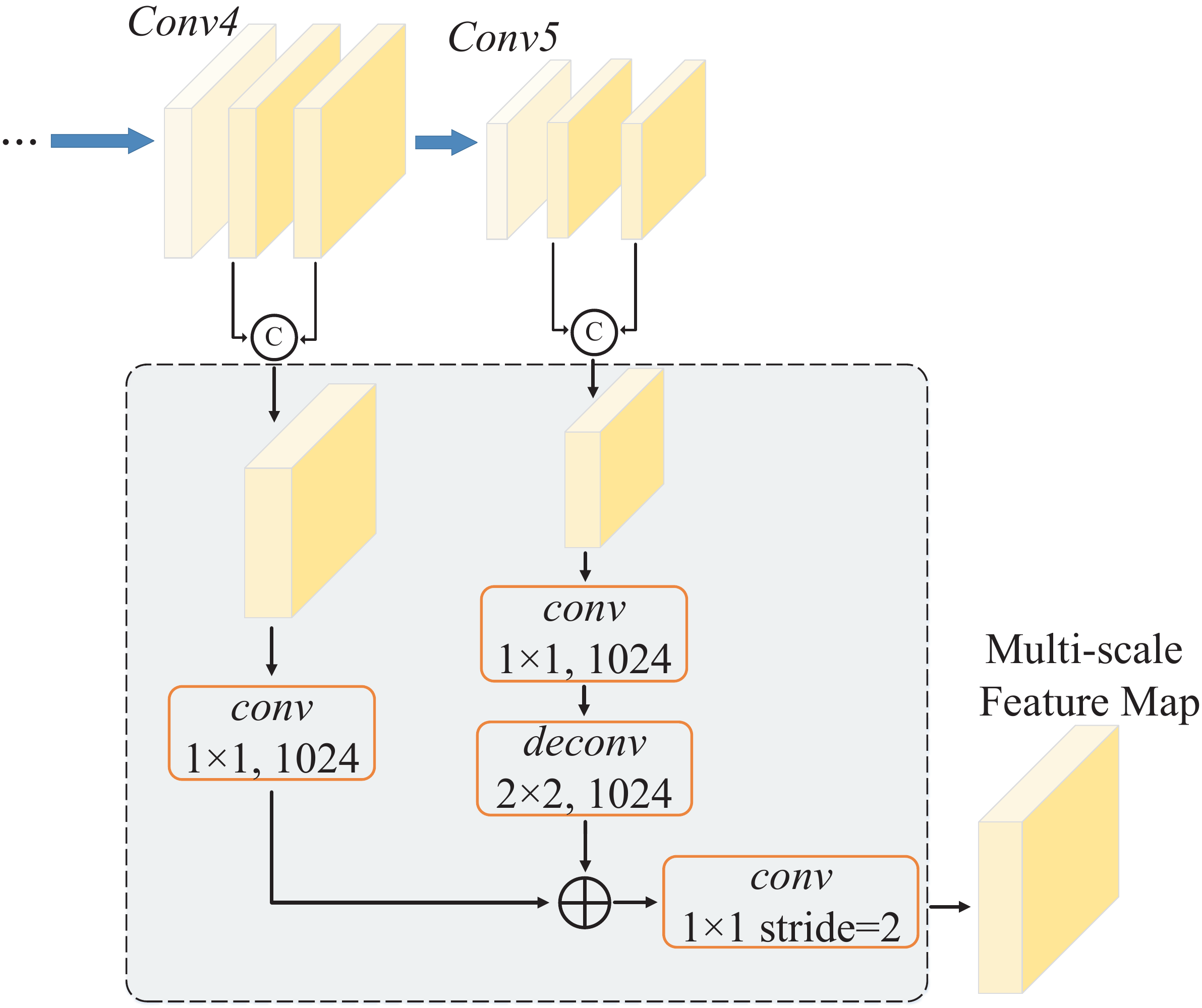}\\
  \vspace{0.1cm}
  \caption{The overview of our modified multi-scale feature map with VGG-16~\cite{simonyan2014very}. Please refer to Fig.~\ref{fig2}.}\label{fig3}
\end{center}\vspace{-.7cm}
\end{figure}

Different from the aforementioned multi-scale strategies, this paper suggests to exploit feature maps of multiple layers in each convolution block. As illustrated in Fig.~\ref{fig3}, our method first concatenates different convolution layers in the same convolution block (\eg, layers of $conv5\_3$ and $conv5\_2$ in VGG-16~\cite{simonyan2014very}), then performs element-wise sum for feature maps of different convolution blocks (\eg, blocks of $conv4$ and $conv5$ for VGG-16~\cite{simonyan2014very}).

Since different convolution blocks have different sizes of feature maps. Feature maps need to share the same size, so that element-wise sum operation can be performed. As suggested in~\cite{shrivastava2016beyond}, an upsampling operation is used to increase the size of feature map in later layer. To this end, a deconvolution layer~\cite{fu2017dssd} is used to enlarge the resolution of feature map in later layer. Finally, we add a $1\times1$ convolution layer with stride 2 for recovering size of feature map in the original model~\cite{ren2015faster}, because the size of feature map has been enlarged two times than original one after upsampling. The experiment results in Sec.~\ref{sec4.2} show our modified multi-scale feature map can achieve higher accuracy with less computational cost.

\subsection{Location-aware Kernel Representation}\label{sec3.2}

The recent progress of challenging fine-grained visual categorization task demonstrates integration of high-order representations with deep CNNs can bring promising improvements~\cite{Cai_2017_ICCV,cui2017kernel,lin2015bilinear,Wang_2017_CVPR}. However, all these methods can not be directly adopted to object detection due to high dimension and missing location information of the feature map. Hence, we present a location-aware polynomial kernel representation to overcome above limitations and to integrate high-order representations into object detection.

Let ${ \mathcal{X} } \in \mathbb{R}^{K \times M \times N}$ be a 3D feature map from a specific convolution layer.
Then, we define a linear predictor $\mathcal{W}$~\cite{Cai_2017_ICCV} on the high-order statistics of $\mathcal{X}$,\vspace{-0.15cm}
\begin{equation}\label{eq1}
f(\mathcal{X}) = \left\langle \mathcal{W}, \sum_{\mathbf{x} \in \mathcal{X}} \phi(\mathbf{x}) \right\rangle
\end{equation}\vspace{-0.15cm}
where $\mathbf{x} \in \mathbb{R}^{K}$ denotes a local descriptor from $\mathcal{X}$. Here, $\sum_{\mathbf{x} \in \mathcal{X}} \phi(\mathbf{x})$ denotes the high-order statistics characterized by a homogenous polynomial kernel~\cite{pham2013fast}.
Thus, Eqn.~\eqref{eq1} can be reformulated as,\vspace{-0.15cm}
%
%Eqn.~\eqref{eq1} can be extended to high-order formulation as\vspace{-0.05cm}:
%\begin{equation}\label{eq2}
%f(x) = \sum_{k=1}^Kw_kx_k+\sum_{r=2}^R\sum_{k_1,\dots,k_r}\mathcal W_{k_1,\dots,k_r}^r(\prod_{s=1}^rx_{k_s})
%\end{equation}\vspace{-0.05cm}
%
\begin{equation}\label{eq3}
f(\mathcal{X}) \!=\! \sum_{\mathbf{x} \in \mathcal{X}} \left\{ \left\langle \mathbf{w}^1,  \mathbf{x} \right\rangle + \sum_{r=2}^R \sum_{k_1,\ldots,k_r}\mathcal W_{k_1,\ldots,k_r}^rx_{k_1}\ldots x_{k_r} \right\}
\end{equation}
where $R$ is the number of order, $\mathcal W^r$ is a $r$-th order tensor containing the weight of order-$r$ predictor, and $x_{k_j}$ denotes the ${k_j}$-th element of $\mathbf{x}$.
%
%It is well known that a $r$-th order tensor can be approximated by a homogenous polynomial kernel~\cite{pham2013fast} as $k(x_1,x_2)=\left\langle\phi(x_1),\phi(x_2)\right\rangle$ and we can transform it to the $r$-th order formation\vspace{-0.05cm}:
%\begin{equation}\label{eq3}
%f(x) = \left\langle w,x\right\rangle+\sum_{r=2}^R\sum_{k_1,\ldots,k_r}\mathcal W_{k_1,\ldots,k_r}^rx_{k_1}\ldots x_{k_r}
%\end{equation}%\vspace{-0.05cm}
%
%
Suppose that $\mathcal W^r$ can be approximated by $D^r$ rank-1 tensors~\cite{kolda2009tensor}, \ie $\mathcal{ W}^r = \sum_{d=1}^{D^r} a^{r,d} \mathbf{u}_1^{r,d} \otimes \ldots \otimes \mathbf{u}_r^{r,d}$.
And Eqn.~\eqref{eq1} can be further rewritten as,\vspace{-0.15cm}
%
%$\mathcal W_{k_1,\ldots,k_r}^r$ is a $r$-th order tensor determining a degree-$r$ homogenous polynomial predictor. Through the rank-1 decomposition~\cite{kolda2009tensor} of the tensor \ie $\mathcal W^r=\sum_{d=1}^D a^{r,d} u^{r,1}\cdot\ldots\cdot u^{r,d}$, Eqn.~\eqref{eq1} can be reformulated for an end-to-end learning:
%
\begin{eqnarray}\label{eq4}
f(\mathcal{X}) &=& \sum_{\mathbf{x} \in \mathcal{X}} \left\{ \left\langle \mathbf{w}^1, \mathbf{x} \right\rangle + \sum_{r=2}^R  \sum_{d=1}^{D^r}a^{r,d} \prod_{s=1}^r \left\langle \mathbf{u}_s^{r,d}, \mathbf{x} \right\rangle \right\}\nonumber  \\
    %&=&  \left\langle w,x\right\rangle +\sum_{r=2}^R\prod_{s=1}^r\sum_{d=1}^{D^r}a^{r,d}\left\langle u_s^{r,d},x_s\right\rangle\nonumber \\
    &=&  \left\{  \left\langle \mathbf{w}^1, \sum_{\mathbf{x} \in \mathcal{X}} \mathbf{x} \right\rangle + \sum_{r=2}^R \left\langle \mathbf{a}^r, \sum_{\mathbf{z}^r \in \mathcal{Z}^r} \mathbf{z}^r \right\rangle \right\}
\end{eqnarray}
where $\mathbf{z}^r = [{z}^{r,1}, \cdots, {z}^{r,D^r}]^\top$ with ${z}^{r,d} = \prod_{s=1}^r \left\langle \mathbf{u}_s^{r,d}, \mathbf{x} \right\rangle$, and $\mathbf{a}^r$ is the weight vector.

Based on Eqn.~\eqref{eq4}, we can compute arbitrary order of representation by learning parameters of weight $\mathbf{w}^1$, $\mathbf{a}^r$ and $\mathbf{u}_s^{r,d}$ ($r = 2, \cdots, R$, $d = 1, \cdots, D^r$, and $s = 1, \cdots, r$).
As suggested in~\cite{Cai_2017_ICCV}, we first focus on the computation of $\mathbf{z}^r$.
Let $\mathcal{Z}^r = \{\mathbf{z}^r\}$.
By defining ${z}^{r}_s = \{\left\langle \mathbf{u}_s^{r,d}, \mathbf{x} \right\rangle\}_{d=1}^{D^r}$, we can then obtain $\mathcal{Z}^{r}_s = \{\mathbf{z}^{r}_s\}$ by performing $r$-th $1\times1$ convolution layers with $D^r$ channels, where $r$ and $D^r$ indicate the number of order and the rank of the tensor, respectively.
In general, $D^r$ is much larger than dimension of original feature map to make a better approximation.
In this way, the high-order polynomial kernel representation can be computed as follows. Given an input feature map $\mathcal X$, we compute feature map ${\mathcal Z^r_s}$ of $r$-th order presentation with performing $r$-th $1\times1$ convolutions with $D^r$ channels on $\mathcal X$ (denoted as
%$\mathcal Z^r_s=[\mathcal Z^{r,1}_s,\dots,\mathcal Z^{r,D^r}_s]$ and
$\mathcal Z^r_s=conv^{r,s}_{1\times1\times D^r}(\mathcal X)$), following by an element-wise product of all feature maps \ie ${\mathcal Z^r}=\mathcal Z^r_1\odot\dots\odot\mathcal Z^r_r$.
Finally, global sum-pooling is adopted to obtain the orderless representation as the input to the linear predictor~\cite{Cai_2017_ICCV}.

The orderless representation, however, is unsuitable for object detection.
The location information is discarded with the introduction of global sum-pooling, thereby making it ineffective to bounding box regression.
Fortunately, we note that $1\times1$ convolution and element-wise product can preserve location information.
Thus, we can simply remove the global sum-pooling operation, and use ${\mathcal Z^r}$ as the kernel representation.
Moreover, the dimension of ${\mathcal Z^r}$ is $D^r \times w\times h$ (\eg, $D^r$ equals to 4,096 in our network), which is far less than the feature map size $c^{2}$ adopted in~\cite{Li_2017_ICCV}, where $c$ is dimension of original feature map $\mathcal X$ (\eg, $c$ equals to 1,024).

\begin{figure}[!tp]
\begin{center}
  \includegraphics[height=4cm]{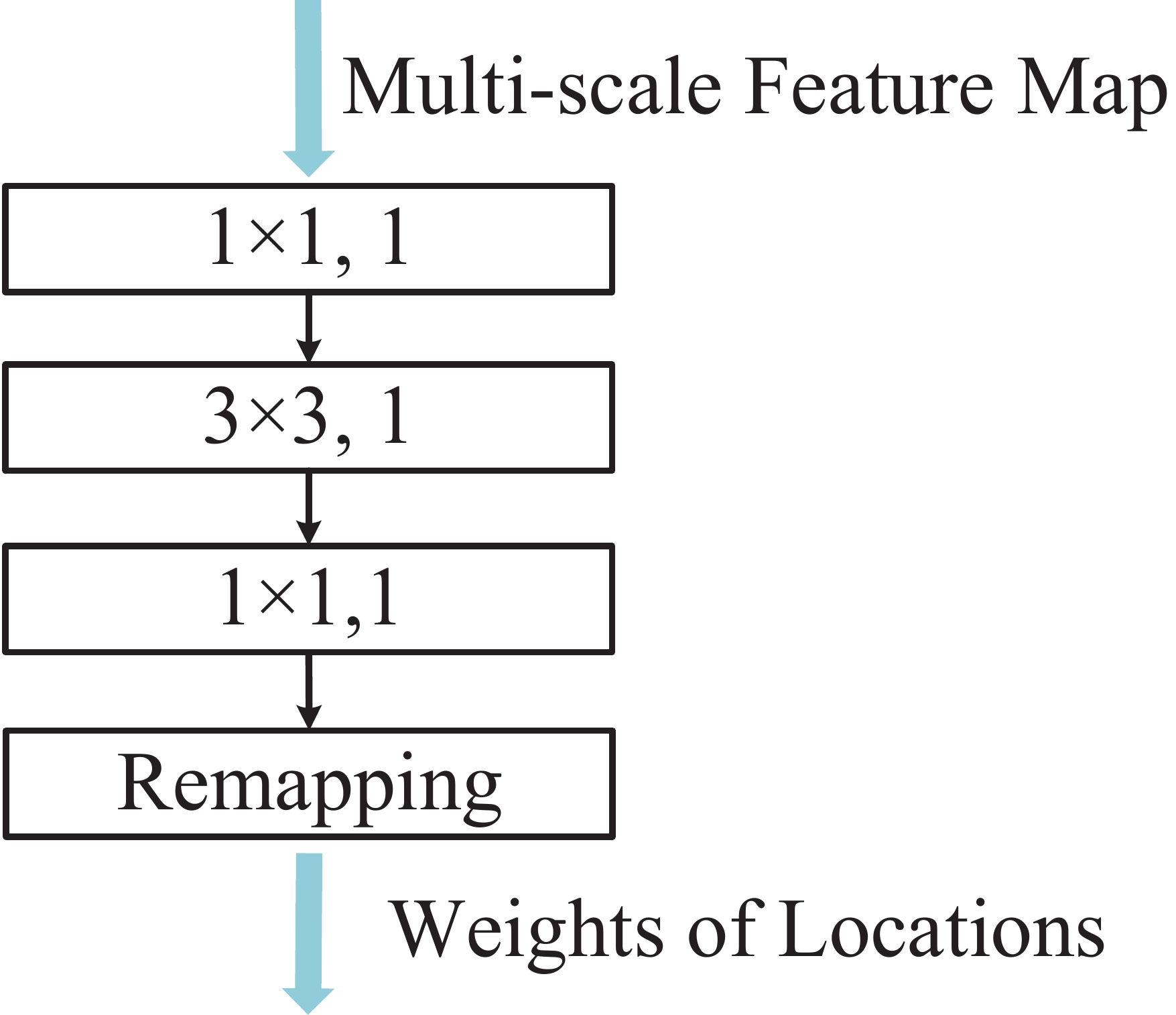}\\
  \vspace{0.1cm}
  \caption{Illustration of proposed location-weight network.}\label{fig4}
\end{center}\vspace{-.7cm}
\end{figure}

Furthermore, parts of the feature maps are more useful to locate objects, and they should be endowed with larger weights.
To this end, we propose a location-aware representation by integrating location weight into the high-order kernel representation. For computing our location-aware kernel representation, we introduce a learnable weight to ${\mathcal Z^r}$,\vspace{-0.15cm}
\begin{equation}\label{eq5}
g_{r}(\mathcal X)=\mathcal Z^r\odot({{\bf 1} \otimes {\mathbf m}(\mathcal X,\Theta_m)}),
\end{equation}\vspace{-0.15cm}
where $\odot$ denotes the element-wise product, $\mathbf m$ is a learnable CNN block with parameter $\Theta_m$ to obtain a location-aware weight and $\bf 1\otimes$ is a re-mapping operation indicating the duplication of matrix $\mathbf{m}$ along channel direction to form a tensor with the same size as $\mathcal Z^r$ for subsequent element-wise product operation as shown in Fig.~\ref{fig4}. A residual block without identity skip-connection is used as our location-weight network. After passing through three different convolutional layers, a weighted feature map is obtained and each point among the feature map represents the contributions to the detection results.

Finally, the representations of $\mathcal X$ with different orders $g_{r}(\mathcal X)_{r=2,\dots,R}$ are concatenated into one single feature map to generate the high-order polynomial kernel representation,\vspace{-0.15cm}
\begin{equation}\label{eq6}
\mathcal{G}(\mathcal X)=[\mathcal X,g_{2}(\mathcal X),\dots,g_{r}(\mathcal X)]^{\top}.
\end{equation}\vspace{-0.15cm}
Moreover, different from using globally average pooling~\cite{Cai_2017_ICCV} to compute polynomial kernel representation, we propose to exploit max RoI pooling on feature map $\mathcal{G}(\mathcal X)$, which computes high-order polynomial kernel representation for each object proposal to preserve location information.

For training our method in an end-to-end manner, the derivatives of $\mathbf{x}$, parameters $\mathbf{u}^{r,d}_s$ and $\Theta_m$ associated with loss function $\mathcal L$ can be achieved according to Eqns.~\eqref{eq4} and~\eqref{eq5},\vspace{-0.15cm}
\begin{equation}\label{eq7}
\frac{\partial\mathcal L}{\partial \mathbf{x}}=\frac{\partial\mathcal L}{\partial g_{r}}(\frac{\partial \mathcal Z^r}{\partial \mathbf{x}}+\sum_{d=1}^{D^r}\mathcal Z^r\frac{\partial {\mathbf m}}{\partial \mathbf{x}})
\end{equation}\vspace{-0.15cm}
\begin{equation}\label{eq8}
\frac{\partial\mathcal L}{\partial \mathbf{u}_s^{r,d}}=\frac{\partial\mathcal L}{\partial g_{r}}\frac{\partial \mathcal Z^r}{\partial \mathbf{u}_s^{r,d}}\odot({{\bf 1}\otimes {\mathbf m}(\mathbf{x},\Theta_m)})
\end{equation}\vspace{-0.15cm}
\begin{equation}\label{eq9}
\frac{\partial\mathcal L}{\partial \Theta_m}=\frac{\partial\mathcal L}{\partial g_{r}}\mathcal Z^r \sum_{d=1}^{D^r}\frac{\partial {\mathbf m}}{\partial \Theta_m}
\end{equation}\vspace{-0.15cm}
where $\frac{\partial\mathcal Z^r}{\partial \mathbf{x}} = \sum_{s=1}^{r}  \sum_{d=1}^{D^r} \prod_{t=1,t\neq s}^r\left\langle \mathbf{u}_t^{r,d}, \mathbf{x} \right\rangle \mathbf{u}_s^{r,d}$, $\frac{\partial \mathcal Z^r}{\partial \mathbf{u}_s^{r,d}}=\prod_{t=1,t\neq s}^r \left\langle \mathbf{u}_t^{r,d}, \mathbf{x} \right\rangle \mathbf{x}$, and $\frac{\partial \mathbf{m}}{\partial \mathbf{x}}$, $\frac{\partial \mathbf{m}}{\partial \Theta_m}$ can be obtained during the back-propagation of the location-weight networks. Although location weight can be learned for different orders of kernel representations by using multiple location-weight networks, we share the same location weight for orders of kernel representations to make a balance between effectiveness and efficiency.

\subsection{MLKP for Object Detection}\label{sec3.3}

\begin{figure}[!t]
\begin{center}
  \includegraphics[width=8.4cm,height=3.cm]{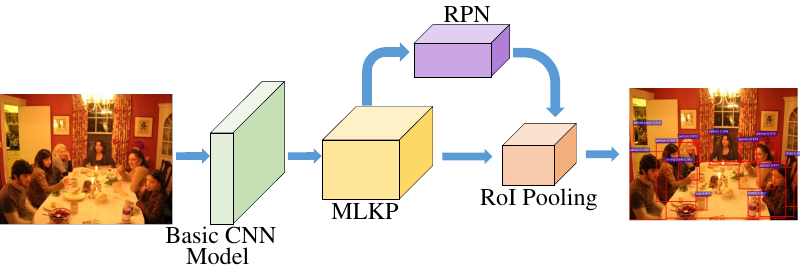}\\
  \vspace{0.1cm}
  \caption{The network of object detection with the proposed MLKP.}\label{fig5}
\end{center}\vspace{-.6cm}
\end{figure}

The above describes the proposed multi-scale location-aware kernel representation (MLKP), and then we illustrate how to apply our  MLKP to object detection. As shown in Fig.~\ref{fig5}, we adopt the similar detection pipeline with Faster R-CNN~\cite{ren2015faster}. Specifically, we first pass an input image through the convolution layers in a basic CNN model (\eg, VGG-16~\cite{simonyan2014very} or ResNet~\cite{he2016deep}). Then, we compute the proposed MLKP on the outputs of convolutional layers while generating object proposals with a region proposal network (RPN). Finally, a RoI pooling layer combining MLKP with RPN is used for classification and regression. This network can be trained in an end-to-end manner.

\section{Experiments}

In this section, we evaluate our proposed method on three widely used benchmarks: PASCAL VOC 2007, PASCAL VOC 2012~\cite{everingham2010pascal} and MS COCO~\cite{lin2014microsoft}. We first describe implementation details of our MLKP, and then make ablation studies on PASCAL VOC 2007. Finally, comparisons with state-of-the-arts on three benchmarks are given.

\subsection{Implementation Details}

To implement our MLKP, as described in Section~\ref{sec3.3}, we employ the similar framework with Faster R-CNN and train our network following existing methods~\cite{Kong_Tao_2017_CVPR,liu2016ssd,kong2016hypernet,dai2016r,fu2017dssd,bell2016inside}. We initialize the network using an ImageNet pretrained basic model~\cite{simonyan2014very,he2016deep} and the weights of layers in MLKP are initialized with the method of ¡±Xavier¡±~\cite{glorot2010understanding}. In the first step we freeze all layers in basic model with only training the layers of MLKP and RPN. Secondly, the whole network is trained within two stages by decreasing learning rate. Our programs are implemented by Caffe Toolkit~\cite{jia2014caffe} on a NVidia 1080Ti GPU. Following the common used settings in Faster R-CNN~\cite{ren2015faster}, the input images are firstly normalized and then we employ two kinds of deep CNNs as basic networks including VGG-16~\cite{simonyan2014very} and RseNet-101\cite{he2016deep}. The mean Average Precision (mAP) is used to measure different detectors. Note that we use single-scale training and testing throughout all experiments, and compare with state-of-the-art methods on three datasets without bells and whistles.

\subsection{Ablation Studies on PASCAL VOC 2007}\label{sec4.2}

In this subsection, we first evaluate the key components of our MLKP on PASCAL VOC 2007, including multi-scale feature map, location-aware polynomial kernel representation, and effect of location-weight network. As suggested in~\cite{ren2015faster}, we train the network on the union set of train/validation in VOC 2007 and VOC 2012, and report the results on test set of VOC 2007 for comparison.

\vspace{-0.45cm}
\paragraph{Multi-scale Feature Map.} In this part, we investigate the effect of different multi-scale feature maps generation strategies on detection performance. In general, integration of more convolutional feature maps brings more improvements, but leads more computational costs. The existing multi-scale methods all only consider feature maps of the last layer in each convolution block. Different from them, we propose a modified strategy to exploit feature maps of multiple layers in each convolution block. Tab.~\ref{tab1} lists the mAP and inference time (Frames Per Second, FPS) of multi-scale feature maps with different convolutional layers on PASCAL VOC 2007. Note that we employ original Faster R-CNN for object detection, aiming at investigating the effect of different strategies.

From the results of Tab.~\ref{tab1}, we can see that integration of more convolutional feature maps indeed brings more improvements. However, feature map with $conv5\_3+conv4\_3+conv3\_3$ obtains only $0.1\%$ gain over one with $conv5\_3+conv4\_3$, but runs about two times slower. For our modified strategy, $conv5\_3/2$ is superior to single layer of $conv5\_3$ over $2.8\%$ with comparable inference time. The gains of two layers in $conv5$ over one may owe to the complementarity of different layers within a convolution block, which enhances representation of proposals. Meanwhile, $conv5\_3/2+conv4\_3/2$ can further improve $conv5\_3/2$ over $0.5\%$ with less additional computational cost. Finally, $conv5\_3/2+conv4\_3/2$ outperforms $conv5\_3+conv4\_3+conv3\_3$ and $conv5\_3+conv4\_3$ with same or less inference time. The above results demonstrate that our modified strategy is more efficient and effective than existing ones.

\begin{table}[!t]
\begin{center}
\begin{tabular}{p{3.25cm}<{\centering}|p{0.6cm}<{\centering}|p{3.1cm}<{\centering}}
\hline
Method & mAP & Inference Time (FPS)\\
\hline
{\small\emph{conv5\_3}} &  73.2 & 15\\
{\small\emph{conv5\_3+conv4\_3}} & 76.2 & 11\\
\small\emph{conv5\_3+conv4\_3+conv3\_3} & 76.3 & 6\\
\hline
{\small\emph{conv5\_3/2}} &  76.0 & 14\\
{\small\emph{conv5\_3/2+conv4\_3/2}} & \bf{76.5} & \bf{11}\\
\hline
\end{tabular}\vspace{0.2cm}
\caption{Comparison of mAP and inference time of multi-scale feature maps with various convolutional layers on PASCAL VOC 2007.}\label{tab1}
\end{center}\vspace{-.5cm}
\end{table}

\begin{table}[!t]
\begin{center}
\begin{tabular}{p{1.15cm}<{\centering}|p{1.7cm}<{\centering}|p{1.2cm}<{\centering}|p{2.55cm}<{\centering}}%{c<{\centering}|c<{\centering}|c<{\centering}c<{\centering}|}
\hline
\multirow{2}{*}{\small Order} & \multirow{2}{*}{\small Dimension}  & \multicolumn{2}{c}{\small {mAP / Inference Time(FPS)}}\\
\cline{3-4}
& & {\small\emph{conv5\_3}} & {\small\emph{conv5\_3/2+conv4\_3/2}}\\
\hline
\small 1 & - & \small {73.2 / 15}& \small {76.5 / 11}\\
\hline
\multirow{2}{*}{2} & \small 2048 & \small {76.4 / 14} & \small {77.7 / 10}\\
\cline{2-4}
& \small 4096 & \small {76.5 / 14} & \small {77.5 / 10}\\
\hline
\multirow{3}{*}{3} & \small 2048 & \small {76.6 / 13}& \small {77.8 /10}\\
\cline{2-4}
& \small 4096 & \small\bf {76.6 / 12}& \small\bf {78.1 / 10}\\
\cline{2-4}
& \small 8192 & \small {76.2 /10}& \small {77.7 / 8}\\
\hline
\end{tabular}\vspace{0.2cm}
\caption{Results of our MLKP with various order-$r$ and dimension $D^r$ under settings of single-scale and multi-scale feature maps.}\label{tab2}
\end{center}\vspace{-.7cm}
\end{table}

\vspace{-0.5cm}
\paragraph{Location-aware Kernel Representation.} Next, we analyze the proposed location-aware kernel representation under settings of single-scale and multi-scale feature maps. As shown in Eqn.~\eqref{eq4}, our location-aware kernel representation involves two parameters, \ie, the order-$r$ and the dimension $D^r$. To obtain compact representations for efficient detection, this paper only considers order-$r=1,2,3$. Meanwhile, the dimension of $D^r$ varies from 2048 to 8192. We summarize the results of MLKP with various order-$r$ and dimension $D^r$ in Tab.~\ref{tab2}, which can be concluded as follows.

\begin{table*}[!t]
\begin{center}
\begin{tabular}{p{2.1cm}<{\centering}|p{0.75cm}<{\centering}|p{0.4cm}<{\centering}|p{0.19cm}<{\centering}p{0.21cm}<{\centering}p{0.2cm}<{\centering}p{0.2cm}<{\centering}p{0.29cm}<{\centering}p{0.19cm}<{\centering}p{0.18cm}<{\centering}p{0.19cm}<{\centering}p{0.24cm}<{\centering}p{0.21cm}<{\centering}p{0.25cm}<{\centering}p{0.20cm}<{\centering}p{0.24cm}<{\centering}p{0.25cm}<{\centering}p{0.26cm}<{\centering}p{0.25cm}<{\centering}p{0.23cm}<{\centering}p{0.20cm}<{\centering}p{0.23cm}<{\centering}p{0.20cm}<{\centering}}
\scriptsize{Method} & \scriptsize{Data} & \scriptsize{mAP} & \scriptsize{aero}&\scriptsize{bike}&\scriptsize{bird}&\scriptsize{boat}&\scriptsize{bottle}&\scriptsize{bus}&\scriptsize{car}&\scriptsize{cat}&\scriptsize{chair}&\scriptsize{cow}&\scriptsize{table}&\scriptsize{dog}&\scriptsize{horse}&\scriptsize{mbike}&\scriptsize{person}&\scriptsize{plant}&\scriptsize{sheep}&\scriptsize{sofa}&\scriptsize{train}&\scriptsize{tv}\\
\hline
\scriptsize Faster R-CNN~\cite{ren2015faster}& \scriptsize 07+12 & \scriptsize{73.2} & \scriptsize76.5 &\scriptsize79.0 &\scriptsize70.9 &\scriptsize65.5 &\scriptsize52.1 &\scriptsize83.1 &\scriptsize84.7 &\scriptsize86.4 &\scriptsize52.0 &\scriptsize81.9 &\scriptsize65.7 &\scriptsize84.8 &\scriptsize84.6 &\scriptsize75.5 &\scriptsize76.7 &\scriptsize38.8 &\scriptsize73.6 &\scriptsize73.9 &\scriptsize83.0 &\scriptsize72.6\\
\scriptsize HyperNet~\cite{kong2016hypernet}& \scriptsize 07+12 & \scriptsize{76.3} & \scriptsize77.4 &\scriptsize83.3 &\scriptsize75.0 &\scriptsize69.1 &\scriptsize62.4 &\scriptsize83.1 &\scriptsize87.4 &\scriptsize87.4 &\scriptsize57.1 &\scriptsize79.8 &\scriptsize71.4 &\scriptsize85.1 &\scriptsize85.1 &\scriptsize80.0 &\scriptsize79.1 &\scriptsize51.2 &\scriptsize\bf79.1 &\scriptsize75.7 &\scriptsize80.9 &\scriptsize76.5\\
\scriptsize ION-Net~\cite{bell2016inside} & \scriptsize 07+12+s & \scriptsize{76.5} & \scriptsize{79.2}&\scriptsize{79.2}&\scriptsize{77.4}&\scriptsize{69.8}&\scriptsize{55.7}&\scriptsize{85.2}&\scriptsize{84.3}&\scriptsize\bf{89.8}&\scriptsize{57.5}&\scriptsize{78.5}&\scriptsize{73.8}&\scriptsize\bf{87.8}&\scriptsize{85.9}&\scriptsize{81.3}&\scriptsize{75.3}&\scriptsize{49.7}&\scriptsize{76.9}&\scriptsize{74.6}&\scriptsize{85.2}&\scriptsize\bf{82.1}\\
\scriptsize SSD300~\cite{liu2016ssd}&\scriptsize 07+12&\scriptsize{77.5} & \scriptsize79.5 &\scriptsize\bf83.9 &\scriptsize76.0 &\scriptsize69.6 &\scriptsize50.5 &\scriptsize\bf87.0 &\scriptsize85.7 &\scriptsize88.1 &\scriptsize60.3 &\scriptsize81.5 &\scriptsize\bf77.0 &\scriptsize86.1 &\scriptsize\bf87.5 &\scriptsize\bf83.9 &\scriptsize79.4 &\scriptsize52.3 &\scriptsize77.9 &\scriptsize\bf79.5 &\scriptsize\bf87.6 &\scriptsize76.8\\
\scriptsize RON384++~\cite{Kong_Tao_2017_CVPR}&\scriptsize 07+12&\scriptsize{77.6} & \scriptsize\bf86.0 &\scriptsize82.5 &\scriptsize76.9 &\scriptsize69.1 &\scriptsize59.2 &\scriptsize86.2 &\scriptsize85.5 &\scriptsize87.2 &\scriptsize59.9 &\scriptsize81.4 &\scriptsize73.3 &\scriptsize85.9 &\scriptsize86.8 &\scriptsize82.2 &\scriptsize\bf79.6 &\scriptsize52.4 &\scriptsize78.2 &\scriptsize76.0 &\scriptsize86.2 &\scriptsize78.0\\
\scriptsize MLKP (Ours)& \scriptsize 07+12 & {\textcolor[rgb]{1,0,0}{\scriptsize\bf78.1}} & \scriptsize78.7 &\scriptsize83.1 &\scriptsize\bf78.8 &\scriptsize\bf71.3 &{\textcolor[rgb]{0,0,1}{\bf\scriptsize{64.4}}} &\scriptsize86.1 &\scriptsize\bf88.0 &\scriptsize87.8 &\scriptsize\bf64.6 &\scriptsize\bf83.2 &\scriptsize73.6 &\scriptsize85.7 &\scriptsize86.4 &\scriptsize81.9 &\scriptsize79.3 &{\textcolor[rgb]{0,0,1}{\bf\scriptsize{53.1}}} &\scriptsize77.2 &\scriptsize76.7 &\scriptsize85.0 &\scriptsize76.1\\
\hline
\scriptsize Faster R-CNN~\cite{ren2015faster}*& \scriptsize 07+12 & \scriptsize{76.4} & \scriptsize79.8 &\scriptsize80.7 &\scriptsize76.2 &\scriptsize68.3 &\scriptsize55.9 &\scriptsize85.1 &\scriptsize85.3 &\scriptsize\bf89.8 &\scriptsize56.7 &\scriptsize87.8 &\scriptsize69.4 &\scriptsize88.3 &\scriptsize88.9 &\scriptsize80.9 &\scriptsize78.4 &\scriptsize41.7 &\scriptsize78.6 &\scriptsize79.8 &\scriptsize85.3 &\scriptsize72.0\\
\scriptsize SSD321~\cite{liu2016ssd}*& \scriptsize 07+12 & \scriptsize{77.1} & \scriptsize76.3 &\scriptsize84.6 &\scriptsize79.3 &\scriptsize64.6 &\scriptsize47.2 &\scriptsize85.4 &\scriptsize84.0 &\scriptsize88.8 &\scriptsize60.1 &\scriptsize82.6 &\scriptsize76.9 &\scriptsize86.7 &\scriptsize87.2 &\scriptsize85.4 &\scriptsize79.1 &\scriptsize50.8 &\scriptsize77.2 &\scriptsize82.6 &\scriptsize\bf87.3 &\scriptsize76.6\\
\scriptsize DSSD321~\cite{fu2017dssd}*& \scriptsize 07+12 & \scriptsize{78.6} & \scriptsize81.9 &\scriptsize84.9 &\scriptsize80.5 &\scriptsize68.4 &\scriptsize53.9 &\scriptsize85.6 &\scriptsize86.2 &\scriptsize88.9 &\scriptsize61.1 &\scriptsize83.5 &\scriptsize\bf78.7 &\scriptsize86.7 &\scriptsize88.7 &\scriptsize\bf86.7 &\scriptsize79.7 &\scriptsize51.7 &\scriptsize78.0 &\scriptsize80.9 &\scriptsize87.2 &\scriptsize79.4\\
\scriptsize R-FCN~\cite{dai2016r}*& \scriptsize 07+12 & \scriptsize{80.5} & \scriptsize79.9 &\scriptsize\bf87.2 &\scriptsize\bf81.5 &\scriptsize72.0 &\scriptsize69.8 &\scriptsize86.8 &\scriptsize\bf88.5 &\scriptsize\bf89.8 &\scriptsize67.0 &\scriptsize\bf88.1 &\scriptsize74.5 &\scriptsize\bf89.8 &\scriptsize\bf90.6 &\scriptsize79.9 &\scriptsize81.2 &\scriptsize53.7 &\scriptsize81.8 &\scriptsize81.5 &\scriptsize85.9 &\scriptsize\bf79.9\\
\scriptsize MLKP (Ours)*& \scriptsize 07+12 & {\textcolor[rgb]{1,0,0}{\scriptsize\bf{80.6}}} & \scriptsize\bf82.2 &\scriptsize83.2 &\scriptsize79.5 &\scriptsize\bf72.9 &{\textcolor[rgb]{0,0,1}{\bf\scriptsize{70.5}}} &\scriptsize\bf87.1 &\scriptsize88.2 &\scriptsize88.8 &\scriptsize\bf68.3 &\scriptsize86.3 &\scriptsize74.5 &\scriptsize88.8 &\scriptsize88.7 &\scriptsize82.0 &\scriptsize\bf81.6 & {\textcolor[rgb]{0,0,1}{\bf\scriptsize{56.3}}} &\scriptsize\bf84.2 &\scriptsize\bf83.3 &\scriptsize85.3 &\scriptsize79.7\\
\end{tabular}\vspace{0.2cm}
\caption{Comparison of different state-of-the-art methods on PASCAL VOC 2007. The methods with/without mark $*$ mean using VGG-16 and ResNet-101 as basic networks, respectively.  "s" indicates that additional segmentation labels are used to train networks.}\label{tab4}
\end{center}\vspace{-.6cm}
\end{table*}

Firstly, our location-aware kernel representation improves the baseline with first-order representation by $3.4\%$ and $1.6\%$  under settings of single and multi-scale feature maps, respectively. It demonstrates the effectiveness of high-order statistics brought by location-aware kernel representation. Secondly, appropriate high-order statistics can achieve promising performance, but the gains tend to saturate as number of order becoming larger. So, integration of overmuch high-order statistics will get fewer gains, and higher dimension $D^r$ mapping does not necessarily lead to better results. Finally, both the multi-scale feature map and location-aware kernel representation can both significantly improve detection performance with less additional inference time, and combination of them achieves further improvement.

\begin{figure}[!t]
\begin{center}
  \includegraphics[width=0.45\textwidth]{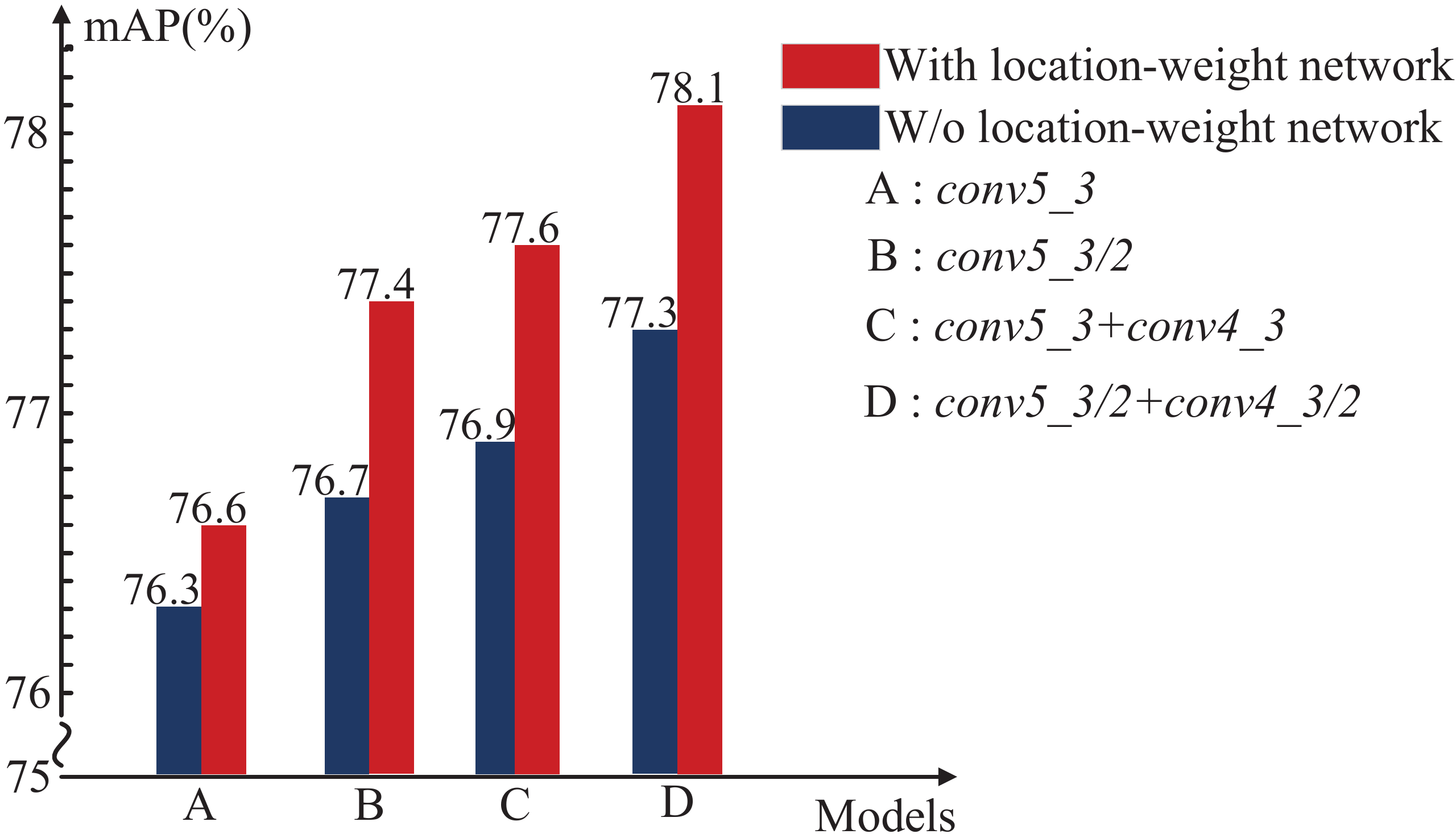}\\
  \caption{Effect of location-aware weight on our MLKP with various feature maps.}\label{fig6}
\end{center}\vspace{-1.0cm}
\end{figure}

\vspace{-0.45cm}
\paragraph{Effect of Location-weight Network.} In the final part of this subsection, we assess the effect of location-weight network on our MLKP. Here, we employ kernel representation with order of 3 and dimension of 4096, which achieves the best result as shown above. Note that our location-weight network in Fig.~\ref{fig4} is very tiny and only cost $\sim7ms$ per image. The results of kernel representation with/without location-weight network are illustrated in Fig.~\ref{fig6}, we can see that location-weight network can achieve $0.3\%\sim0.8\%$ improvement under various settings of feature maps. Meanwhile, for more effective feature maps, location-weight network obtains bigger gains. Note when more effective feature maps of $conv5\_3/2+conv4\_3/2$ are employed, location-weight network obtains $0.8\%$ improvement, which is nontrivial since the counterpart without location-weight network gets a strong baseline result ($77.3\%$).\vspace{-0.05cm}

%\begin{table}[!htp]
%\begin{center}
%\begin{tabular}{p{3.7cm}<{\centering}|p{1.5cm}<{\centering}}
%\hline
%Method & mAP \\
%\hline
%conv5\_3 &  76.6\\
%conv5\_3/2 &  77.4\\
%conv5\_3+conv4/3 & 77.6\\
%conv5\_3/2+conv4\_3/2 & 78.1\\
%\hline
%\end{tabular}
%\end{center}
%\caption{Effect of location-aware weight on our MKLP under various feature maps.}\label{tab3}
%\end{table}

\subsection{Comparison with State-of-the-arts}\label{section4.3}

To further evaluate our method, we compare our MLKP with several recently proposed state-of-the-art methods on three widely used benchmarks: \ie, PASCAL VOC 2007, PASCAL VOC 2012~\cite{everingham2010pascal} and MS COCO~\cite{lin2014microsoft}.\vspace{-0.25cm}

\subsubsection{Results on PASCAL VOC 2007}\vspace{-0.05cm}

On PASCAL VOC 2007, we compare our method with seven state-of-the-art methods. Specifically, the network is trained for 120k iterations for the first step with learning rate of 0.001. Then, whole network is fine-tuned for 50k iterations with learning rate of 0.001 and 30k iterations with learning rate of 0.0001. The weight decay and momentum are set to 0.0005 and 0.9 in the total three steps, respectively. For a fair comparison, all competing methods except R-FCN~\cite{dai2016r} utilize single input size without multi-scale training/testing and box voting. The results (mAP and AP of each category) of different methods with VGG-16~\cite{simonyan2014very} or ResNet-101~\cite{he2016deep} models are listed in Tab.~\ref{tab4}.

\begin{table*}[!t]
\begin{center}
\begin{tabular}{p{2.1cm}<{\centering}|p{0.75cm}<{\centering}|p{0.35cm}<{\centering}|p{0.19cm}<{\centering}p{0.21cm}<{\centering}p{0.21cm}<{\centering}p{0.21cm}<{\centering}p{0.25cm}<{\centering}p{0.19cm}<{\centering}p{0.18cm}<{\centering}p{0.19cm}<{\centering}p{0.24cm}<{\centering}p{0.21cm}<{\centering}p{0.25cm}<{\centering}p{0.21cm}<{\centering}p{0.25cm}<{\centering}p{0.26cm}<{\centering}p{0.27cm}<{\centering}p{0.25cm}<{\centering}p{0.25cm}<{\centering}p{0.20cm}<{\centering}p{0.25cm}<{\centering}p{0.20cm}<{\centering}}
\scriptsize{Method} & \scriptsize{Data} & \scriptsize{mAP} & \scriptsize{aero}&\scriptsize{bike}&\scriptsize{bird}&\scriptsize{boat}&\scriptsize{bottle}&\scriptsize{bus}&\scriptsize{car}&\scriptsize{cat}&\scriptsize{chair}&\scriptsize{cow}&\scriptsize{table}&\scriptsize{dog}&\scriptsize{horse}&\scriptsize{mbike}&\scriptsize{person}&\scriptsize{plant}&\scriptsize{sheep}&\scriptsize{sofa}&\scriptsize{train}&\scriptsize{tv}\\
\hline
\scriptsize Faster R-CNN~\cite{ren2015faster}&\scriptsize 07++12&\scriptsize{70.4} & \scriptsize84.9 &\scriptsize79.8 &\scriptsize74.3 &\scriptsize53.9 &\scriptsize49.8 &\scriptsize77.5 &\scriptsize75.9 &\scriptsize88.5 &\scriptsize45.6 &\scriptsize77.1 &\scriptsize55.3 &\scriptsize86.9 &\scriptsize81.7 &\scriptsize80.9 &\scriptsize79.6 &\scriptsize40.1 &\scriptsize72.6 &\scriptsize60.9 &\scriptsize81.2 &\scriptsize61.5\\
\scriptsize HyperNet~\cite{kong2016hypernet}&\scriptsize 07++12&\scriptsize{71.4} & \scriptsize84.2 &\scriptsize78.5 &\scriptsize73.5 &\scriptsize55.6 &\scriptsize53.7 &\scriptsize78.7 &\scriptsize79.8 &\scriptsize87.7 &\scriptsize49.6 &\scriptsize74.9 &\scriptsize52.1 &\scriptsize86.0 &\scriptsize81.7 &\scriptsize83.3 &\scriptsize81.8 &\scriptsize48.6 &\scriptsize73.5 &\scriptsize59.4 &\scriptsize79.9 &\scriptsize65.7\\
\scriptsize SSD512~\cite{liu2016ssd}&\scriptsize 07++12&\scriptsize{74.9} & \scriptsize\bf87.4 &\scriptsize82.3 &\scriptsize75.8 &\scriptsize59.0 &\scriptsize52.6 &\scriptsize\bf81.7 &\scriptsize\bf81.5 &\scriptsize90.0 &\scriptsize55.4 &\scriptsize79.0 &\scriptsize59.8 &\scriptsize88.4 &\scriptsize84.3 &\scriptsize84.7 &\scriptsize83.3 &\scriptsize50.2 &\scriptsize78.0 &\scriptsize66.3 &\scriptsize\bf86.3 &\scriptsize\bf72.0\\
\scriptsize RON384++~\cite{Kong_Tao_2017_CVPR}&\scriptsize 07++12&\scriptsize{75.4} & \scriptsize86.5 &\scriptsize82.9 &\scriptsize76.6 &\scriptsize\bf60.9 &\scriptsize55.8 &\scriptsize\bf81.7 &\scriptsize80.2 &\scriptsize91.1 &\scriptsize\bf57.3 &\scriptsize\bf81.1 &\scriptsize\bf60.4 &\scriptsize87.2 &\scriptsize\bf84.8 &\scriptsize\bf84.9 &\scriptsize81.7 &\scriptsize51.9 &\scriptsize\bf79.1 &\scriptsize\bf68.6 &\scriptsize84.1 &\scriptsize70.3\\
\scriptsize MLKP (Ours)& \scriptsize 07++12 & {\textcolor[rgb]{1,0,0}{\scriptsize\bf{75.5}}} & \scriptsize86.4 &\scriptsize\bf83.4 &\scriptsize\bf78.2 &\scriptsize60.5 &{\textcolor[rgb]{0,0,1}{\bf\scriptsize{57.9}}} &\scriptsize80.6 &\scriptsize79.5 &\scriptsize\bf91.2 &\scriptsize56.4 &\scriptsize81.0 &\scriptsize58.6 &\scriptsize\bf91.3 &\scriptsize84.4 &\scriptsize84.3 &\scriptsize\bf83.5 &{\textcolor[rgb]{0,0,1}{\bf\scriptsize{56.5}}} &\scriptsize77.8 &\scriptsize67.5 &\scriptsize83.9 &\scriptsize67.4\\
\hline
\scriptsize Faster R-CNN~\cite{ren2015faster}*& \scriptsize 07++12 & \scriptsize{73.8} & \scriptsize86.5 &\scriptsize81.6 &\scriptsize77.2 &\scriptsize58.0 &\scriptsize51.0 &\scriptsize78.6 &\scriptsize76.6 &\scriptsize\bf93.2 &\scriptsize48.6 &\scriptsize80.4 &\scriptsize59.0 &\scriptsize92.1 &\scriptsize85.3 &\scriptsize84.8 &\scriptsize80.7 &\scriptsize48.1 &\scriptsize77.3 &\scriptsize66.5 &\scriptsize84.7 &\scriptsize65.6\\
\scriptsize SSD321~\cite{liu2016ssd}*& \scriptsize 07++12 & \scriptsize{75.4} & \scriptsize\bf87.9 &\scriptsize82.9 &\scriptsize73.7 &\scriptsize61.5 &\scriptsize45.3 &\scriptsize81.4 &\scriptsize75.6 &\scriptsize92.6 &\scriptsize57.4 &\scriptsize78.3 &\scriptsize\bf65.0 &\scriptsize90.8 &\scriptsize\bf86.8 &\scriptsize85.8 &\scriptsize81.5 &\scriptsize50.3 &\scriptsize78.1 &\scriptsize75.3 &\scriptsize\bf85.2 &\scriptsize72.5\\
\scriptsize DSSD321~\cite{fu2017dssd}*& \scriptsize 07++12 & \scriptsize{76.3} & \scriptsize87.3 &\scriptsize83.3 &\scriptsize75.4 &\scriptsize\bf64.6 &\scriptsize46.8 &\scriptsize\bf82.7 &\scriptsize76.5 &\scriptsize92.9 &\scriptsize\bf59.5 &\scriptsize78.3 &\scriptsize64.3 &\scriptsize91.5 &\scriptsize86.6 &\scriptsize\bf86.6 &\scriptsize82.1 &\scriptsize53.3 &\scriptsize79.6 &\scriptsize\bf75.7 &\scriptsize\bf85.2 &\scriptsize\bf73.9\\
\scriptsize MLKP(Ours)*& \scriptsize 07++12 & {\textcolor[rgb]{1,0,0}{\scriptsize\bf{77.2}}} & \scriptsize87.1 &\scriptsize\bf85.1 &\scriptsize\bf79.0 &\scriptsize64.2 & {\textcolor[rgb]{0,0,1}{\scriptsize\bf{60.3}}} &\scriptsize82.1 &\scriptsize\bf80.6 &\scriptsize92.3 &\scriptsize57.4 &\scriptsize\bf81.8 &\scriptsize61.6 &\scriptsize\bf92.1 &\scriptsize86.3 &\scriptsize85.3 &\scriptsize\bf84.3 & {\textcolor[rgb]{0,0,1}{\bf\scriptsize{59.1}}} &\scriptsize\bf81.7 &\scriptsize69.5 &\scriptsize85.0 &\scriptsize70.1\\
\end{tabular}\vspace{0.2cm}
\caption{Comparison of different state-of-the-art methods on PASCAL VOC 2012. The methods with/without mark $*$ mean using VGG-16 and ResNet-101 as basic networks, respectively. Our result can be found at http:\url{//host.robots.ox.ac.uk:8080/anonymous/TENHEH.html}, \url{ http://host.robots.ox.ac.uk:8080/anonymous/XI6YFV.html}.}\label{tab5}
\end{center}\vspace{-.55cm}
\end{table*}

\begin{table*}[!t]
\begin{center}
\begin{tabular}{c<{\centering}|c<{\centering}|c<{\centering}c<{\centering}c<{\centering}|c<{\centering}c<{\centering}c<{\centering}|c<{\centering}c<{\centering}c<{\centering}|c<{\centering}c<{\centering}c<{\centering}c<{\centering}}
\multirow{2}{*}{\footnotesize Method} & \multirow{2}{*}{\footnotesize Training set}  & \multicolumn{3}{c|}{\footnotesize Avg.Precision, IOU:} & \multicolumn{3}{c|}{\footnotesize Avg.Precision, Area:} & \multicolumn{3}{c|}{\footnotesize Avg.Recall, \#Det:} & \multicolumn{3}{c}{\footnotesize Avg.Recall, Area:}\\
\cline{3-14}
& &   \footnotesize0.5:0.95 & \footnotesize0.50 & \footnotesize0.75 & \footnotesize Small & \footnotesize Med. & \footnotesize Large & \footnotesize 1 &  \footnotesize10 & \footnotesize 100 & \footnotesize Small & \footnotesize Med. & \footnotesize Large\\
\hline
\footnotesize Faster R-CNN~\cite{ren2015faster} &\footnotesize trainval  & \footnotesize21.9 & \footnotesize42.7 & \footnotesize23.0 & \footnotesize6.7 & \footnotesize25.2 & \footnotesize34.6 & \footnotesize22.5 & \footnotesize32.7 & \footnotesize33.4 & \footnotesize10.0 & \footnotesize38.1 & \footnotesize53.4\\
\footnotesize ION~\cite{bell2016inside} &\footnotesize train+s  & \footnotesize24.9 & \footnotesize44.7 & \footnotesize25.3 & \footnotesize7.0 & \footnotesize26.1 & \footnotesize40.1 & \footnotesize23.9 & \footnotesize33.5 & \footnotesize34.1 & \footnotesize10.7 & \footnotesize 38.8 & \footnotesize54.1\\
\footnotesize SSD300~\cite{liu2016ssd} &\footnotesize trainval35  & \footnotesize25.1 & \footnotesize43.1 & \footnotesize25.8 & \footnotesize6.6 & \footnotesize25.9 & \footnotesize41.4 & \footnotesize23.7 & \footnotesize35.1 & \footnotesize37.2 & \footnotesize11.2 & \footnotesize 40.4 & \footnotesize58.4\\
\footnotesize SSD512~\cite{liu2016ssd} &\footnotesize trainval35  & \footnotesize26.8 & \footnotesize46.5 & \footnotesize\bf27.8 & \footnotesize\bf9.0 & \footnotesize28.9 & \footnotesize\bf41.9 & \footnotesize24.8 & \footnotesize37.5 & \footnotesize\bf39.8 & \footnotesize14.0 & \footnotesize 43.5 & \footnotesize59.0\\
\footnotesize MLKP (Ours) &\footnotesize trainval35  & \bf\footnotesize{\color{red}26.9} & \bf\footnotesize48.4 & \footnotesize26.9 & \footnotesize8.6 & \bf\footnotesize29.2 & \footnotesize41.1 & \bf\footnotesize25.6 & \bf\footnotesize37.9 & \footnotesize38.9 & \bf\footnotesize16.0 & \bf\footnotesize 44.1 & \bf\footnotesize59.0\\
\hline
%\footnotesize Faster R-CNN*~\cite{he2016deep} &\footnotesize train  & \footnotesize27.2 & \footnotesize41.5 & - & - & - & - & - & - & - & - & - & -\\
\footnotesize DSSD321~\cite{fu2017dssd}* &\footnotesize trainval35  & \footnotesize28.0 & \footnotesize45.4 & \footnotesize29.3 & \footnotesize6.2 & \footnotesize28.3 & \footnotesize\bf49.3 & \footnotesize25.9 & \footnotesize37.8 & \footnotesize39.9 & \footnotesize11.5 & \footnotesize 43.3 & \footnotesize\bf64.9\\
\footnotesize SSD321~\cite{fu2017dssd}* &\footnotesize trainval35  & \footnotesize28.0 & \footnotesize46.1 & \footnotesize29.2 & \footnotesize7.4 & \footnotesize28.1 & \footnotesize47.6 & \footnotesize25.5 & \footnotesize37.1 & \footnotesize39.4 & \footnotesize12.7 & \footnotesize 42.0 & \footnotesize62.6\\
%\footnotesize R-FCN~\cite{dai2016r}* &\footnotesize trainval  & \footnotesize29.2 & \footnotesize51.5 & - & \footnotesize10.3 & \footnotesize32.4 & \footnotesize45.0 & - & - & - & - & - & -\\
\footnotesize MLKP (Ours)* &\footnotesize trainval35  & \bf\footnotesize{\color{red}28.6} & \bf\footnotesize52.4 & \bf\footnotesize31.6 & \footnotesize\bf10.8 & \bf\footnotesize33.4 & \footnotesize45.1 & \bf\footnotesize27.0 & \bf\footnotesize40.9 & \footnotesize\bf41.4 & \footnotesize\bf15.8 & \bf\footnotesize 47.8 & \footnotesize62.2\\
%\footnotesize R-FCN &\footnotesize trainval  & \footnotesize29.2 & \footnotesize51.5 & - & \footnotesize10.3 & \footnotesize32.4 & \footnotesize43.3 & - & - & - & - & - & -\\
\end{tabular}\vspace{0.2cm}
\caption{Comparison of different state-of-the-art methods on MS COCO {\emph{2017test-dev}}. "s" indicates training networks with additional segmentation labels. The methods with/without mark $*$ mean using VGG-16 and ResNet-101 as base networks, respectively.}\label{tab6}
\end{center}\vspace{-.6cm}
\end{table*}

As reported in Tab.~\ref{tab4}, when VGG-16 model is employed, our MLKP improves Faster R-CNN by $4.9\%$. Because of sharing the exactly similar framework with Faster R-CNN, we owe the significant gains to the proposed MLKP. Meanwhile, our method also outperforms HyperNet~\cite{kong2016hypernet}, ION-Net~\cite{bell2016inside} and RON~\cite{Kong_Tao_2017_CVPR} over $1.8\%$, $1.6\%$ and $0.5\%$, respectively. The improvements over aforementioned methods demonstrate the effectiveness of location-aware high-order kernel representation. In addition, our MLKP is superior to state-of-the-art region-free method SSD~\cite{liu2016ssd} by $0.6\%$.

Then we adopt ResNet-101 model, our MLKP outperforms Faster R-CNN by $4.2\%$. Meanwhile, it is superior to DSSD~\cite{fu2017dssd} which incorporates multi-scale information into SSD method~\cite{liu2016ssd}. Additionally, MLKP slightly outperforms R-FCN~\cite{dai2016r}, even R-FCN exploits multi-scale training/testing strategy. Note that R-FCN obtains only mAP of $79.5\%$ using single-scale training/testing.\vspace{-0.3cm}

\subsubsection{Results on PASCAL VOC 2012}

Following the same experimental settings on PASCAL VOC 2007, we compare our method with five state-of-the-art methods on PASCAL VOC 2012. We train network on training and validation sets of PASCAL VOC 2007 and PASCAL VOC 2012 with additional test set of PASCAL VOC 2007. The results of different methods on test set of VOC 2012 are reported in Tab.~\ref{tab5}. Our MLKP achieves the best results with both VGG-16 and ResNet-101 models, and it improves Faster R-CNN over $5.1\%$ and $3.4\%$, respectively. These results verify the effectiveness of our MLKP again. Note that the results on both VOC 2007 and 2012 show our method can achieve impressive improvement in detecting small and hard objects, \eg, bottles and plants.\vspace{-0.25cm}

\subsubsection{Results on MS COCO}\vspace{-0.05cm}

Finally, we compare our MLKP with four state-of-the-art methods on the challenging MS COCO benchmark~\cite{lin2014microsoft}. MS COCO contains 80k training images, 40k validation images and 80k testing images from 80 classes. We train our network on trainval35 set following the common settings~\cite{liu2016ssd} and report the results getting from \emph{test-dev2017} evaluation sever.
%\footnote{\url{https://competitions.codalab.org/competitions/5181}}.
Because \emph{test-dev2017} and \emph{test-dev2015} contain the same images, so the results obtaining from them are comparable. As MS COCO is a large-scale benchmark, we employ four GPUs to accelerate the training process. We adopt the same settings of hyper-parameters with PASCAL VOC datasets, but train the network with more iterations as MS COCO containing much more training images. Specifically, we train the network in the first step with 600k iterations, and performs fine-tuning with learning rate of 0.001 and 0.0001 for 150k and 100k iterations, respectively. We adopt the single-train and single-test strategy, and use the standard evaluating metric on MS COCO for comparison.

The comparison results are given in Tab.~\ref{tab6}. As Faster R-CNN only reported two numerical results, we conduct additional experiments using the released model in ~\cite{ren2015faster}.  When adopting VGG-16 network, our MLKP can improve the Faster R-CNN over $5.0\%$ at IOU=[0.5:0.05:0.95], and is superior to other competing methods. For the ResNet-101 backbone, our method outperforms state-of-the-art region-free methods DSSD and SSD by $0.6\%$. In particular, the proposed MLKP improves competing methods in detecting small or medium size objects.

\section{Conclusion}

This paper proposes a novel location-aware kernel approximation method to represent object proposals for effective object detection, which, to our best knowledge, is among the first which exploits high-order statistics in improving performance of object detection. Our MLKP takes full advantage of statistical information of multi-scale convolution features. The significant improvement over the first-order counterparts demonstrates the effectiveness of our MLKP. The experimental results on three popular benchmarks show our method is very competitive, indicating integration of high order statistics is a encouraging direction to improve object detection. As our MLKP is framework-independent, we plan to extend it other frameworks.

{\small
\bibliographystyle{ieee}
\bibliography{egbib}
}

\end{document}